\documentclass[10pt,twocolumn,letterpaper]{article}

\usepackage{cvpr}              

\definecolor{cvprblue}{rgb}{0.21,0.49,0.74}
\usepackage[pagebackref,breaklinks,colorlinks,allcolors=cvprblue]{hyperref}

\def\paperID{*****} 
\def\confName{CVPR}
\def\confYear{2025}

\title{Design of an Expression Recognition Solution Based on the Global Channel-Spatial Attention Mechanism and Proportional Criterion Fusion}
\author{
Jun Yu$^1$, Yang Zheng$^1$, Lei Wang$^1$\thanks{Corresponding author}, Yongqi Wang$^1$, Shengfan Xu$^2$\\
$^1$University of Science and Technology of China\\
$^2$Macau University of Science and Technology\\
\tt\small \{harryjun,wangl\}@ustc.edu.cn\\
\tt\small \{zhengyang,wangyongqi\}@mail.ustc.edu.cn \\
\tt\small \{eyki1010\}@163.com \\
}

\begin{document}
\maketitle
\begin{abstract}

Facial expression recognition is a challenging classification task that holds broad application prospects in the field of human-computer interaction. This paper aims to introduce the method we will adopt in the 8th Affective and Behavioral Analysis in the Wild (ABAW) Competition, which will be held during the Conference on Computer Vision and Pattern Recognition (CVPR) in 2025.
First of all, we apply the frequency masking technique and the method of extracting data at equal time intervals to conduct targeted processing on the original videos. Then, based on the residual hybrid convolutional neural network and the multi-branch convolutional neural network respectively, we design feature extraction models for image and audio sequences. In particular, we propose a global channel-spatial attention mechanism to enhance the features initially extracted from both the audio and image modalities respectively.
Finally, we adopt a decision fusion strategy based on the proportional criterion to fuse the classification results of the two single modalities, obtain an emotion probability vector, and output the final emotional classification. We also design a coarse - fine granularity loss function to optimize the performance of the entire network, which effectively improves the accuracy of facial expression recognition.
In the facial expression recognition task of the 8th ABAW Competition, our method ranked third on the official validation set. This result fully confirms the effectiveness and competitiveness of the method we have proposed. 
\end{abstract}    
\section{Introduction}

The problem of facial expression recognition involves multiple research fields such as artificial intelligence, bioscience, and psychology, and it is a popular research branch within the field of artificial intelligence. Traditional facial expression recognition mainly focuses on the research of a single modality of facial images. However, the emotional information covered by it is limited, and different modalities of data also have different sensitivities to emotional states. Therefore, integrating multiple modalities has become a research trend. There are mainly three strategies for the integration of multiple modalities: feature-level fusion, decision-level fusion, and model fusion. Among them, the decision-level fusion method does not require strict temporal synchronization between modalities, and it also solves the problem of different feature reliabilities of different modalities. Therefore, it is a popular choice in modern emotion recognition architectures. 

Researchers conduct modeling and analysis of human emotions through audio - act data\cite{5} such as audio, text, and facial images, as well as physiological signal data such as electroencephalogram (EEG) and body temperature\cite{1}. Among them, the two modalities of audio and facial images account for up to 93\%\cite{3} of the conveyed emotional factors. The two are usually complementary and can provide relatively comprehensive information for the emotion recognition task. In recent years, deep - learning - related algorithms such as Convolutional Neural Network (CNN), Residual Network (ResNet), Long Short - Term Memory (LSTM)\cite{9}, and attention mechanisms have been widely applied to the research of audio signal feature extraction and audio emotion classification. For example, Kumaran\cite{1} et al. proposed a deep convolutional recurrent neural network model; Chen\cite{5} et al. proposed a Bidirectional Long and Short Term Memory (BiLSTM) network based on the attention mechanism. However, currently, dual - modality fusion models still have problems such as complex network structures, a large number of model parameters, and difficulty in training.

In order to promote interdisciplinary cooperation and address key research issues spanning affective computing, machine learning, and multimodal signal processing, Kollias et al. took the lead in launching the "Affective Behavior Analysis in the Wild" (ABAW) initiative\cite{70}. The 8th ABAW Workshop and Competition is scheduled to be held in conjunction with the IEEE CVPR conference in 2025.  Aiming at the above emotion recognition problems, this paper conducts research based on audio and static facial images, and the main contributions are as follows:

1.We propose a channel - spatial attention mechanism suitable for audio - visual emotion recognition. By combining the channel attention, channel shuffling, and spatial attention mechanisms, this mechanism aims to capture the global dependencies in the feature maps and enhance the feature representation ability of the extracted original audio and images.

2.We use a proportional decision fusion strategy to obtain the final emotion classification results. A coarse - fine granularity loss function is designed to optimize the performance of the entire network.

3.In the 8th ABAW Competition, our algorithm achieved excellent results on the official validation set. This result fully confirms the effectiveness and competitiveness of the proposed method. 
\section{Related Work}
\subsection{Image-based facial expression recognition}

Various visual descriptors can be extracted from facial morphology to detect expressions in video streams\cite{73,79,80,84}. For example, Nguyen et al.\cite{29} extracted 68 facial landmark points and constructed 32 geometric descriptors to train a Support Vector Machine (SVM) classifier for differentiating different emotional categories. The framework proposed in reference is based on Facial Action Units (AUs), which continuously detects emotional states through AUs. With the development of deep learning technology, visual - based emotion recognition systems using 2D/3D Convolutional Neural Network (CNN) architectures, taking video frames or sequences as input, have higher recognition rates compared to traditional frame - aggregation - based methods. Two - dimensional convolutional neural networks like the Emotional Deep Attention Network (EmotionalDAN) \cite{31}aim to solve the problems of emotion, valence, and landmark recognition in one step. The Spatial Transformer Network (STN)\cite{32} can detect the main regions of interest in video frames and correct spatial variations, and a similar framework proposed in reference is used to capture facial landmark points or facial visual saliency maps. Facial Expression Recognition (FER) from video streams regards a series of frames within a temporal analysis window as a single input. For example, the C3D network used in reference\cite{35}, which utilizes 3D convolutional kernels with shared weights along the time axis, has been widely applied in dynamic facial expression recognition\cite{39}.There are also methods to design facial expression recognition models trained on single images. Reference\cite{41} proposed supervised and self - supervised learning methods to improve the classification accuracy in fine - grained and in - the - wild scenarios. The Visual Transformer with Feature Fusion (VTFF) in reference\cite{42} can recognize emotions under extreme conditions. The Transformer - based Facial Expression Recognition (TransFER) method in reference\cite{43} extracts rich relation - aware representations between visual descriptors. Reference\cite{44} proposed a Graph Convolutional Network (GCN) framework to exploit the dependencies between two types of emotion recognition tasks. Abbasi et al.\cite{45} constructed a graph - based representation for children's facial expression recognition. Reference\cite{46} introduced a video - based facial expression recognition method to improve the descriptiveness of embedding features using emotion - wheel information. Overall, the analysis of these advanced methods shows that it is important to distinguish between peak and non - peak video frames in video sequences. Although deep Recurrent Neural Networks/3D Convolutional Neural Networks encode the temporal dependencies of consecutive frames, their performance is not satisfactory and they are difficult to train. 

\subsection{Audio - integrated emotion recognition}

A number of studies have applied multimodal audio-visual analysis to emotion prediction. For example, Nguyen et al.\cite{47} integrated 3D-CNN and deep belief networks, and fused the visual and audio feature vectors using the bilinear pooling theory to recognize emotions. In reference\cite{48}, different acoustic and visual features were input into a CNN extended with an RNN, and then an average fusion was carried out at the decision-making level. Kahou et al.\cite{49} integrated multiple deep neural networks for different data modalities to predict emotions.VAANet\cite{50} used a specific CNN architecture combined with a late fusion strategy, but ignored the interaction between features. AVER\cite{51}, Wang et al.\cite{52}, reference\cite{53}, reference\cite{54}, etc. processed multimodal information in different ways respectively. Recently, reference\cite{55} proposed a model that fuses audio-visual features at the model level. Hu et al.\cite{56}, MEmoBERT\cite{57}, reference\cite{60}, and multiple transformer-based frameworks\cite{59,61} have also made their own explorations. Experiments have shown that the performance of multimodal concatenation is better.

Overall, the performance of multimodal emotion recognition is superior to that of unimodal recognition. The model-level fusion in has a good effect\cite{55}. The importance of information in video sequences varies. Most methods adopt a two-stage shallow processing pipeline and do not make use of the complementarity between modalities. Moreover, existing methods do not consider the correlation between emotion categories, such as the polarity defined by Mikel's emotion wheel\cite{62}.
We propose a cross-modal audio-visual fusion framework, which employs the global channel-spatial attention mechanism and the decision fusion strategy based on the proportional criterion. Except for the initial pre-training, no auxiliary data is required. Through operations such as splitting the video stream and extracting audio and images, multiple attention mechanisms are utilized to process features and predict emotions. Meanwhile, a coarse - fine granularity loss function is designed to optimize the performance of the entire network. The next section will provide a detailed introduction to this method and its modules. 

\begin{figure*}[ht]
\centering
\includegraphics[width=\linewidth]{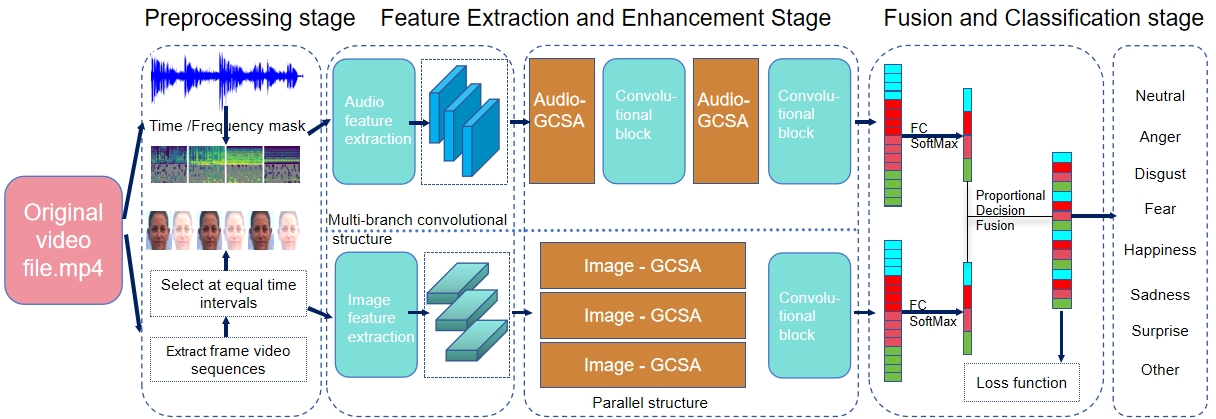}
\caption{Our proposed framework for Expression Recognition.}
\label{Figure 1}
\end{figure*}
\section{Method}

\subsection{Overview}

The overall framework of the audio-visual emotion recognition training network in this paper is shown in Figure 1. The entire network is implemented in an end-to-end manner. According to the processing flow of the task, it can be divided into three stages: data preprocessing, feature extraction and feature enhancement, and modality fusion and classification.
First is the data preprocessing stage. Face image sequences and audio files are extracted from the original audio-visual data. The frequency masking technique is applied to conduct targeted processing on the original audio data. At the same time, image data content is selected at equal time intervals to ensure the representativeness and regularity of the samples.
Then, based on the residual hybrid convolutional neural network and the multi-branch convolutional neural network methods respectively, feature extraction models for face image sequences and audio sequences are designed.
Subsequently, the obtained sequence features are enhanced by using the global channel-spatial attention module.
Finally, a proportional decision fusion strategy is adopted to fuse the classification results of the two single modalities, obtaining an emotion probability vector, and the final emotion classification is output based on this emotion probability vector.
In order to optimize the model more efficiently, a coarse - fine granularity loss function is designed, and the coarse granularity penalty coefficient is determined according to the emotion probability vector. 

\subsection{Feature Extraction}

This paper employs a Multi-branch Convolutional Neural Network (MCNN) and a Residual-based Hybrid Convolutional Neural Network (RHCNN) respectively for the extraction of emotional features from audio and facial images.

As shown in the following formula, the audio features extracted by the MCNN are denoted as $F_a$, and $A$ refers to the input audio sequence. 
against changes in emotional features.
\begin{equation}
F_{a} = \text{MCNN}(A)
\end{equation}
The MCNN structure is composed of branches for extracting features in the local, temporal, and spatial dimensions. Each different branch consists of two convolutional blocks with different receptive field sizes. In this way, receptive fields of various sizes can be utilized to fully obtain the important emotional context information in the feature maps. This not only enriches the input features but also enhances the robustness of the model 

Similarly, the visual features extracted by the RHCNN are denoted as $F_v$, and $V$ refers to the input image sequence.
\begin{equation}
F_{v} = \text{RHCNN}(V)
\end{equation}
The RHCNN structure is implemented based on the concept of residual connections and depthwise separable convolutions. Firstly, the image data is input into two convolutional blocks. Each convolutional block is composed of a convolutional layer with a receptive field size of N×N, a Batch Normalization layer (BN), and a ReLU activation function layer, which are used to extract primary features and increase the nonlinearity of the features. The residual connection block is composed of a 1×1 convolution, an N×N depthwise separable convolution, and a 1×1 convolution. The 1×1 convolution block is used to change the number of channels and increase the nonlinear representation of the network, making it easier for residual connections. The receptive field size of N×N can capture more spatial information, and a channel mixing operation is used subsequently to enhance the information exchange among the channels of the feature maps. 

\subsection{ Global Channel-Spatial Attention}
After completing the preliminary feature extraction, in order to enhance the feature representation ability, we conduct enhancement processing on the extracted audio and visual features. We have designed a Global Channel-Spatial Attention (GCSA) module to enhance the representational ability of the input feature maps. This module combines channel attention, channel shuffling, and spatial attention mechanisms, aiming to capture the global dependencies in the feature maps.

\subsubsection{Channel Attention Sub-module}

In traditional convolutional neural networks, the relationships between channels may be overlooked. However, these relationships are crucial for capturing global information. If the dependencies between channels are not taken into account, the model may fail to fully utilize all the information in the feature maps, resulting in insufficient capture of global features.To enhance the dependencies between channels, we adopt a Multi-Layer Perceptron (MLP) in the channel attention sub-module. Implementing channel attention through the MLP can reduce the redundant information between channels and highlight the important features. First, we permute the dimensions of the input feature map so that the channel dimension is the last one. Then, we process it through a two-layer MLP. In the first layer, the number of channels is reduced to 1/4 of the original. A ReLU activation function is used to introduce non - linearity. In the second layer, the number of channels is restored to the original dimension. In this way, the global dependencies between channels can be better captured. Finally, we perform an inverse permutation to restore the original dimensions and generate a channel attention map through a Sigmoid activation function. The input feature map is multiplied element - by - element with the channel attention map to obtain an enhanced feature map.
\begin{equation}
{F}_{\text{channel}}= \sigma(MLP(\text{Permute}(\mathbf{F}_{\text{input}}))) \odot \mathbf{F}_{\text{input}}
\end{equation}

\subsubsection{Channel Shuffle}

Even after enhancing the channel attention, there may still be a problem that the information between channels is not fully mixed. If the information between channels is not adequately mixed, the feature representation ability may be limited, and the effect of channel attention cannot be fully realized.

To further mix and share information, a channel shuffle operation is applied. The enhanced feature map is divided into 4 groups, with each group containing C/4 channels. A transpose operation is performed on the grouped feature map to disrupt the channel order within each group. Subsequently, the shuffled feature map is restored to its original shape (C×H×W). In this way, the feature information can be better mixed, and the feature representation ability can be enhanced.
\begin{equation}
{F}_{\text{shuffle}} = \text{ChannelShuffle}(\mathbf{F}_{\text{channel}})
\end{equation}

\subsubsection{Spatial Attention Sub - module}

Relying solely on channel attention and channel shuffle operations may not fully utilize spatial information. However, spatial information is equally important for capturing local and global features in an image. If the information in the spatial dimension is not considered, the model may overlook important details in the feature maps.

In the spatial attention sub - module, the input feature map passes through a 7x7 convolutional layer, and the number of channels is reduced to 1/4 of the original. Then, batch normalization and a ReLU activation function are applied for non - linear transformation. Next, a second 7x7 convolutional layer restores the number of channels to the original dimension C, followed by another batch normalization layer. Finally, a Sigmoid activation function is used to generate a spatial attention map. The shuffled feature map is multiplied element - by - element with the spatial attention map to obtain the final output feature map.
\begin{equation}
F_{\text{spatial}} = \sigma(\text{Conv}(\text{BN}(\text{ReLU}(\text{Conv}(F_{\text{shuffle}}))))) \odot F_{\text{shuffle}}
\end{equation}

\subsection{Decision Fusion Based on the Proportion Criterion}
There are mainly two common multimodal fusion methods: feature fusion and decision fusion. The former fuses the features of each modality and then makes an emotion judgment on the fused features. The latter, however, fuses the emotion state judgments of each modality and outputs the final decision. Compared with feature fusion, decision fusion doesn't need to consider the semantic differences between modalities, and it is simpler and more effective. In this paper, a decision fusion strategy based on the proportion criterion is adopted to fuse the audio and image modalities. To obtain the emotion state judgments of audio and image, the audio features and video features are respectively input into the fully - connected layers. Then, the SoftMax function is used to convert them into probability values, thus completing the mapping from features to emotion probability vectors.The emotion probability vectors of audio and image are denoted as $P_a$ and $P_v$ respectively. According to the proportion criterion, the emotion probability vectors of the two modalities are fused to output the probabilities of each discrete emotion. The formula of the proportion criterion is as follows:
\begin{equation}
P = \frac{mP_{a} + nP_{v}}{m + n}
\end{equation}

\noindent
Among them, $P$ is the emotion probability vector of the audio - video modality. $m$ and $n$ are the proportion parameters of the two modalities, where $0 < m < 6$, $0 < n < 6$, and $m\in\mathbb{Z}$, $n\in\mathbb{Z}$. 

\subsection{Coarse - Fine Granularity Loss}
For discrete emotion classification, the cross-entropy loss function is mostly used to train the model. This loss function calculates the relative entropy between the predicted value and the true value based on the probability vector. The predicted value and the true value are measured in terms of the discrete emotion category \(e\), and it can be regarded as the calculation of the fine-grained classification loss that contains a single emotion category. Denote \(P = [p_{i1}, p_{i2}, \ldots, p_{im}]\) as the probability vector, where \(p_{ie}\) is the predicted probability that the data sample \(i\) belongs to the true category \(e\). The cross-entropy loss function is defined as follows: 
\begin{equation}
L = -\frac{1}{N}\sum_{i = 1}^{N}\sum_{e = 1}^{M} y_{ie} \log(p_{ie})
\end{equation}

\noindent
Among them, \(N = 1, 2, \ldots, n\) represents the number of data samples, \(M = 1, 2, \ldots, m\) represents the number of emotion categories. \(y_{ie}\) is an indicator function. When the predicted category of data sample \(i\) is the same as the true category \(e\), \(y_{ie}=1\); otherwise, \(y_{ie} = 0\).

After the discrete emotion categories are coarsely grained, a qualitative analysis of the emotion categories is achieved, while the fine-grained categories are a quantitative division of the emotion categories. Theoretically, a qualitative error in an event can lead to a chain reaction of errors. Therefore, the loss caused by the misjudgment of the coarse-grained emotion categories should be amplified. In response to this, this paper adds a coarse - grained penalty coefficient on the basis of the cross - entropy loss while maintaining the fine - grained classification loss, and designs a coarse and fine granularity loss function, which is shown as follows:
\begin{equation}
L_{cf} = - \frac{1}{N} \sum_{1}^{N} (1 + \mu_{mn}) \sum_{e = 1}^{M} y_{ie} \log(p_{ie})
\end{equation}

\noindent
Among them, $\mu_{mn}$ ($m \neq n$) is the coarse - grained penalty coefficient. Based on the four coarse - grained emotion categories divided by four quadrants, there are 12 situations of coarse - grained misjudgment. To simplify the calculation, the two events of "misjudging $E_m$ as $E_n$" and "misjudging $E_n$ as $E_m$" are merged into "the confusion event between $E_m$ and $E_n$", that is, $\mu_{mn}=\mu_{nm}$. Therefore, this paper defines six coarse - grained penalty coefficients, denoted as $\mu_{12}$, $\mu_{13}$, $\mu_{14}$, $\mu_{23}$, $\mu_{24}$, and $\mu_{34}$, and their values are determined by specific error events. 

\section{Experiment}

In this section, we will provide a detailed description of the used datasets, the experiment setup, and the experimental results.

\subsection{Datasets}

The 8th Workshop and Competition on Affective and Behavioral Analysis in the Wild has announced the Aff-wild2 database, which is a key collection extracted from a series of studies. This resource serves as the core of the EXPR Classification Challenge\cite{71,72,74,75,76,77,78,81,82,83,85}. Its audio-visual dataset consists of 548 videos and approximately 2.7 million frames, which are annotated for six basic facial expressions, as well as the neutral and "other" categories.

To enrich the depth of the dataset, we have integrated data from the AFEW and CK+ databases. The AFEW dataset is a corpus of dynamic temporal facial expression data in a near real-world environment extracted from movies\cite{86}. It contains 1,426 video clips in AVI format, divided into seven discrete emotions. The CK+ dataset provides data from 123 subjects, which were filmed in a laboratory environment\cite{87}. It includes videos of six universal expressions and the neutral expression, with the expression labels being categorical discrete values. This integration aims to enhance the comprehensiveness and practicality of our analysis. By using the Generative Adversarial Network (GAN) technology, new expression video samples are generated based on the data features of AFEW and CK+. These samples conform to the overall data distribution characteristics of Aff-wild2, thereby increasing the diversity and quantity of the data and enhancing the comprehensiveness of our dataset.

\subsection{Setup}

All the experiments in this paper are developed based on Python 3.10 and the PyTorch 2.2.0 deep learning framework. A GPU of the RTX 3090 model with a video memory of 24GB is used to train and test the model. The batch size is set to 16, and the model is iterated for 300 rounds. The Stochastic Gradient Descent (SGD) optimizer is selected, with a learning rate of 0.0001. The learning rate for the weight parameters of audio and facial images is set to 0.001. The Dropout in the fully - connected layer is set to 0.2, and the momentum is set to 0.9. The coarse - fine granularity loss is chosen as the loss function. The smaller the loss, the higher the prediction accuracy of the model.  

\begin{table}[htbp]
\centering
\caption{Ablation study results on the validation set}
\label{tab:Ablation study results on the validation set.}
\begin{tabular}{ccc}
\toprule
\textbf{Method} & \textbf{F1 Score (\%)} & \textbf{Modality} \\ \midrule
Official & 23.00 & - \\
Base & 31.89 & V \\
Base + GCSA & 33.53 & V \\ 
Base + GCSA + CF & 35.59 & V\\
Base & 42.62 & V + A \\
Base + GCSA & 45.92 & V + A \\ 
Base + GCSA + CF & 46.51 & V + A\\
\bottomrule
\end{tabular}
\end{table}

\subsection{Metrics}

In line with the competition requirements, we employ the average F1 score as our evaluation metric, which is robust against class frequency variations and particularly suitable for imbalanced class distributions. The calculation of the average F1 score is as follows:

\begin{equation}
F_{1}^{c} = \frac{2\times \text{Precision} \times \text{Recall}}{\text{Precision} + \text{Recall}}
\end{equation}

\begin{equation}
F1 = \frac{1}{N}\sum_{c = 1}^{N}F_{1}^{c}
\end{equation}

\noindent
where N represents the number of classes and c means c-th class.

\subsection{Results}

According to the official competition results, the method we submitted ranked third. To verify the effectiveness of our method, we conducted ablation studies on each component and strategy within this method. These studies were carried out respectively on single-modal data and dual-modal data, and the results are shown in Table 1.It is evident that the application of our technology significantly improved the recognition performance, and the recognition accuracy was even higher in the dual-modal scenario, which validates that the feature extraction enhancement and the unique fusion strategy can further improve the accuracy. In addition, incorporating the global channel-spatial attention mechanism can increase the accuracy by 3.3\%, highlighting the importance of feature enhancement. Ultimately, through post-processing, the accuracy of the model reached 46.51\%. 
\section{Conclusion}

In this paper, we propose a method of fusing audio feature expressions to improve the ability of facial expression recognition. The first stage is the data preprocessing stage. In the video processing pipeline, the preprocessing stage plays a crucial foundational role. At this stage, the frequency masking technique is first applied to conduct targeted processing on the original audio data, thus filtering and optimizing the information. Subsequently, the image data content is selected at equal time intervals to ensure the representativeness and regularity of the samples.
The second stage is the feature extraction and enhancement stage. To address the issue of insufficient modal representation capabilities, we adopt the global channel-spatial attention mechanism to enhance the features initially extracted from both the audio and image modalities, thereby improving the representation capabilities of these modalities.
Finally, there is the decision-making and classification stage. We adopt a decision fusion strategy based on the proportional principle to obtain the final emotional classification result. A coarse - fine granularity loss function is designed to optimize the performance of the entire network.
Ultimately, we won the third place in the Expression Recognition track of the 8th Affective and Behavioral Analysis in the Wild (ABAW) Competition. In the future, we will continue to explore ways to achieve more accurate dynamic facial expression recognition and classification results. 
{
    \small
    \bibliographystyle{ieeenat_fullname}
    \bibliography{main}
}


\end{document}